\documentclass{article}       

\usepackage[preprint]{nips_2018}

\usepackage[utf8]{inputenc} 
\usepackage[T1]{fontenc}    
\usepackage{hyperref}       
\usepackage{url}            
\usepackage{booktabs}       
\usepackage{amsfonts}       
\usepackage{nicefrac}       
\usepackage{microtype}      

\usepackage{graphicx}     

\def\persistency{persistency}      

\title{Faster SGD training by minibatch \persistency}

\author{
  Matteo Fischetti \thanks{corresponding author: \texttt{http://www.dei.unipd.it/\textasciitilde fisch}} \\
  Department of Information Engineering\\
  University of Padova, Italy\\
  \texttt{matteo.fischetti@unipd.it} \\
  \And
  Iacopo Mandatelli \\
  Department of Information Engineering\\
  University of Padova, Italy\\
  \texttt{iacopo.mandatelli@studenti.unipd.it} \\  
	\And
  Domenico Salvagnin \\
  Department of Information Engineering\\
  University of Padova, Italy\\
  \texttt{domenico.salvagnin@unipd.it} 
}

\begin{document}

\maketitle

\begin{abstract}
It is well known that, for most datasets, the use of large-size minibatches for Stochastic Gradient Descent (SGD) typically leads to slow convergence and poor generalization. On the other hand, large minibatches are of great practical interest as they allow for a better exploitation of modern GPUs. Previous literature on the subject concentrated on how to adjust the main SGD parameters (in particular, the learning rate) when using large minibatches. In this work we introduce an additional feature, that we call \emph{minibatch \persistency}, that consists in reusing the same minibatch for $K$ consecutive SGD iterations. The computational conjecture here is that a large minibatch contains a significant sample of the training set, so one can afford to slightly overfitting it without worsening generalization too much. The approach is intended to speedup SGD convergence, and also has the advantage of reducing the overhead related to data loading on the internal GPU memory. We present computational results on CIFAR-10 with an  AlexNet architecture, showing that even small \persistency\ values ($K=2$ or $5$) already lead to a significantly faster convergence and to a comparable (or even better) generalization than the standard ``disposable minibatch'' approach ($K=1$), in particular when  large minibatches are used.
The lesson learned is that minibatch \persistency\ can be a simple yet effective way to deal with large minibatches.

\end{abstract}

\section{The idea} \label{sec:intro}

Gradient Descent is one of the methods of choice in Machine Learning and, in particular, in Deep Learning \cite{book}. For large training sets, the computation of the true gradient of the (decomposable) loss function is however very expensive, so the \emph{Stochastic Gradient Descent} (SGD) method---in its \emph{on-line} or \emph{minibatch} variant---is highly preferable in practice because of its superior speed of convergence and degree of generalization. 

In recent years, the availability of parallel architectures (notably, GPUs) suggested the use of larger and larger minibatches, which however can be problematic because it is known that the larger the minibatch, the poorer the convergence and generalization properties; see, e.g., \cite{Wilson2003,LeCun2012,Keskar2016,minibatch32}. Thus, finding a practical way of feeding massively-parallel GPUs with large minibatches is nowadays a hot research topic. 

Recent work on the subject pointed out the need of modifying the SGD parameters (in particular, the learning rate) to cope with large minibatches
\cite{Wilson2003,LeCun2012,Keskar2016,minibatch32}, possibly using clever strategies such as learning-rate linear scaling and warm-up \cite{Goyal2017}. All the published works however give for granted that a ``disposable minibatch'' strategy is adopted, namely: within one epoch, the current minibatch is changed at each SGD iteration. 

In this paper we investigate a different strategy that reuses a same minibatch for $K$ (say) consecutive SGD iteration, where parameter $K$ is called \emph{minibatch \persistency} ($K=1$ being the standard rule). Our intuition is that large minibatches contain a lot of information about the training set, that we do not want to waste by dropping them too early. In a sense, we are ``slightly overfitting'' the current minibatch by investing $K-1$ additional SGD iterations on it, before replacing it with a different minibatch. The approach also has the practical advantage of reducing the computational overhead related to the operation of loading new data into the GPU memory.

Of course, insisting too much on a same minibatch is a risky policy in terms of overfitting, hence one has to computationally evaluate the viability of the approach by quantifying the pros and cons of the minibatch-persistency idea on some well-established setting.

In the present paper we report the outcome of our computational tests on the well-known CIFAR-10 \cite{cifar10} dataset, using (a reduced version of) the AlexNet \cite{alexnet} architecture, very much in the spirit of the recent work presented in \cite{minibatch32}. Our results show that small \persistency\ values $K=2$ or 5 already produce a significantly improved performance (in terms of training speed and generalization) with minibatches of size 256 or larger. The lesson learned is that the use of large minibatches becomes much more appealing when combined with minibatch \persistency, at least in the setting we considered.



\section{Experiments} \label{sec:experiments}  

We next report the outcome of some tests we performed to evaluate the practical impact of minibatch persistency on improving training time without affecting generalization in a negative way.

\subsection{Setup} \label{sec:setup}   

As already mentioned, our experimental setting is very similar to the one in \cite{minibatch32}. We addressed the CIFAR-10 \cite{cifar10} dataset with a reduced version of the AlexNet \cite{alexnet} architecture involving convolutional layers with stride equal to 1, kernel sizes equal to [11,5,3,3,3], number of channel per layer equal to [64,192,384,256,256], max-pool layers with 2x2 kernels and stride 2, and 256 hidden nodes for the fully-connected layer. As customary, the dataset was shuffled and partitioned into 50,000 examples for the training set, and the remaining 10,000 for the test set. 

Our optimization method was SGD with momentum \cite{SGDmomentum} with cross entropy loss---a combination of softmax and negative log-likelihood loss (NLLLoss). The gradient of a minibatch of size $m$ was computed as the sum of the $m$ gradients of the given examples in the minibatch, with respect to the considered loss function. To be more specific, let $L_i(\theta)$ denote the contribution to the loss function of the $i$-th training example in the minibatch, with respect to the weight vector $\theta$. For every minibatch of size $m$, the weight update rule at iteration $t$ is as follows: 
\begin{eqnarray*}
	\Delta\theta_t = \sum_{i=1}^{m} \nabla_\theta L_i(\theta_t) \\
	v_t = \gamma v_{t-1} - \mu \: \Delta\theta_t \\
	\theta_{t+1} = \theta_t + v_t .
\end{eqnarray*}                                                                                                         

Training was performed for 100 epochs using PyTorch \cite{pytorch}, with learning rate $\mu=0.001$ and momentum coefficient $\gamma=0.5$. The initial random seed (affecting, in particular, the random weight initialization) was kept the same for all (deterministic) runs.    

All runs have been performed on a PC equipped with a single GPU (Nvidia GTX1080 Ti); reported computing times are in wall-clock seconds.

Our experiments are just aimed at evaluating the impact of different values of the minibatch-persistency parameter $K$ for a given minibatch size $m$. This is why we decided to use the very basic training algorithm described above. In particular, batch normalization \cite{Ioffe2015}, dropout \cite{dropout} and data augmentation were not used in our experiments. In addition, momentum coefficient and learning rate were fixed a priori (independently of the minibatch size and not viewed as hyper-parameters to tune) and used in all the reported experiments; see, e.g., \cite{Wilson2003} for a discussion about the drawbacks of using a learning rate independent of the minibatch size. As a result, the final test-set accuracies we reached after 100 epochs (about 60\%) are definitely not competitive with the state of the art. Of course, better results are expected when using more sophisticated training strategies---some preliminary results in this directions are reported in Section~\ref{sec:conclusions}.

\subsection{Results} \label{sec:results}

\begin{figure} 
  \centering          
  	\small{Test accuracy} 
		\includegraphics[width=1.0\textwidth]{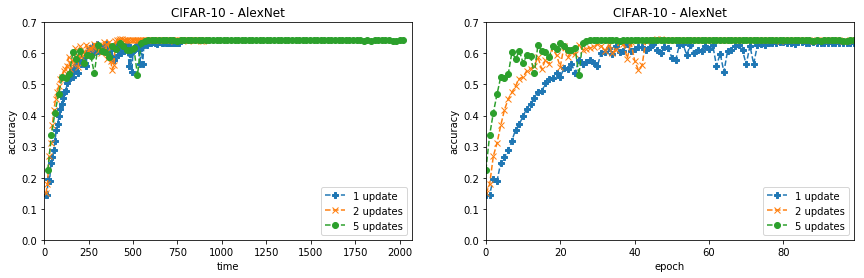}
 
		\small{Test loss} 
		\includegraphics[width=1.0\textwidth]{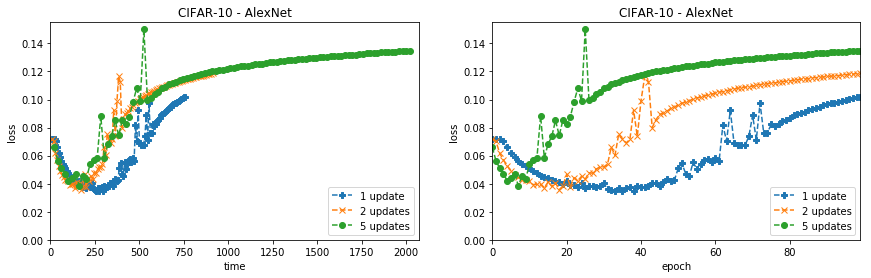}

 	\caption{Results with minibatch size $m=32$ and minibatch \persistency\ $K$ = 1 (standard), 2, and 5.}                 
	\label{fig:alexnet_32}
\end{figure}

\begin{figure} 
  \centering          
  	\small{Test accuracy} 
		\includegraphics[width=1.0\textwidth]{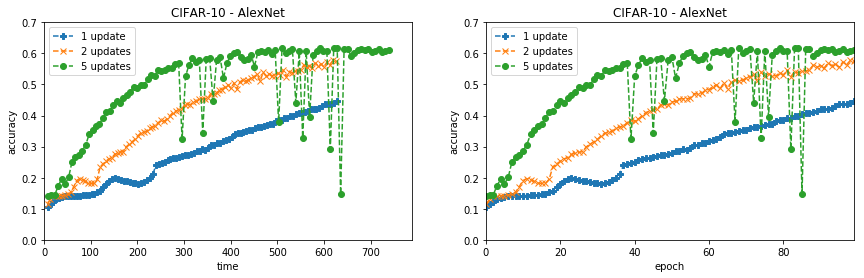}
		\small{Test loss} 
		\includegraphics[width=1.0\textwidth]{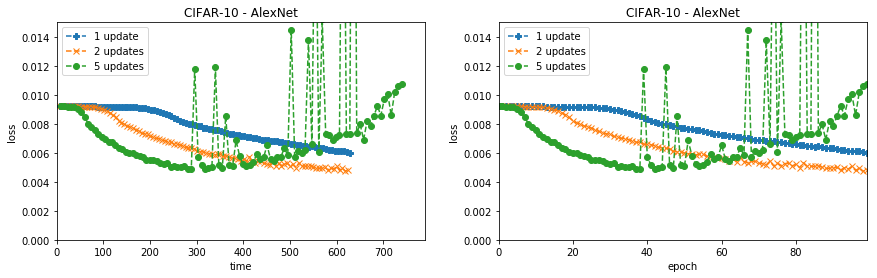}

 	\caption{Results with minibatch size $m=256$ and minibatch \persistency\ $K$ = 1 (standard), 2, and 5.}                 
	\label{fig:alexnet_256}
\end{figure}

\begin{figure} 
  \centering          
  	\small{Test accuracy} 
		\includegraphics[width=1.0\textwidth]{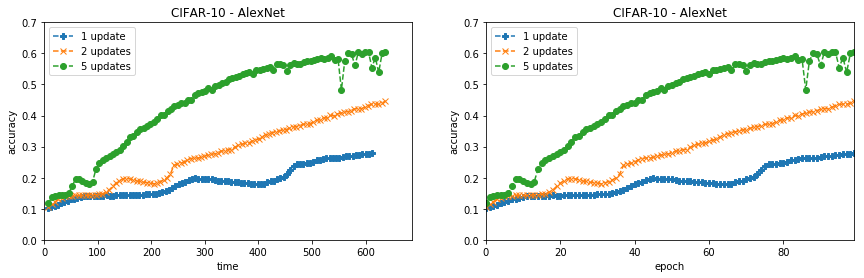}
 
		\small{Test loss} 
		\includegraphics[width=1.0\textwidth]{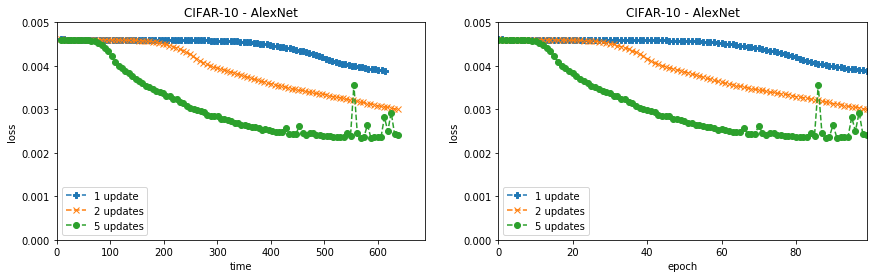}

 	\caption{Results with minibatch size $m=512$ and minibatch \persistency\ $K$ = 1 (standard), 2, and 5.}                 
	\label{fig:alexnet_512}
\end{figure}

Figures \ref{fig:alexnet_32},  \ref{fig:alexnet_256} and \ref{fig:alexnet_512} report the results of our experiments for minibatch sizes $m$ = 32, 256 and 512, respectively. Note that, according to \cite{minibatch32}, the best performance for CIFAR-10 and a (reduced) AlexNet architecture is achieved for $m \le 8$, while minibatches of size $256$ or $512$ are considered too large to produce competitive results in the classical setting---as we will see, this is no longer true when minibatch \persistency\ is used.

Each figure plots top-1 accuracy and loss function (both on the test set) for minibatch-persistency parameter $K\in \{1, 2, 5\}$, as a function of the total computing time (subfigures on the left) and of the number of epochs (on the right). Note that, within a single epoch, each training example is evaluated $K$ times, hence one would expect the computing time to perform each epoch be multiplied by $K$. This is why, to have a fair comparison of different values of $K$, it is important to report computing times explicitly.

Figure \ref{fig:alexnet_32} refers to a small minibatch of size $m=32$. Higher values of $K$ give improved accuracy in terms of epochs (subfigure on the right), however this does not translate into an improvement of computing time (subfigure on the left). Indeed, completing the 100 epochs with $K=5$ took approximately 3 times more time than case $K=1$ (about 2000 instead of 700 sec.s), which is in any case significantly less than the factor of 5 one would expect. On the other hand, the run with $K=5$ reached its best accuracy of $0.64450$ after 30 epochs and about 600 sec.s, while the run with $K=1$ reached its best accuracy of $0.63640$ only at epoch 80, after 600 sec.s as well. As to loss, the runs with larger $K$ started overfitting much earlier, both in terms of epochs and computing times. This behavior is not unexpected, as the minibatch is too small to be representative of the whole training set, and reusing it several times is prone to overfitting.

The overall picture radically changes when a medium-size minibatch with $m=256$ examples is considered (Figure \ref{fig:alexnet_256}). Here, overfitting was less pronounced, and the computing time to complete the 100 epochs did not increase significantly with the value of $K$---thus confirming that minibatch persistency has a positive effect in terms of GPU exploitation when the minibatch is not too small.   

The positive effect of minibatch persistency is even more evident when larger minibatches come into play (case $m=512$ of Figure \ref{fig:alexnet_512}). Here, the run with $K=5$ is the clear winner, both in terms of accuracy and loss.  

The above results are rather encouraging, and show that the use of large minibatches becomes much more appealing when combined with minibatch \persistency.

\newpage

\begin{figure} 
  \centering          
  	\small{Test accuracy} 
 		\includegraphics[width=1.0\textwidth]{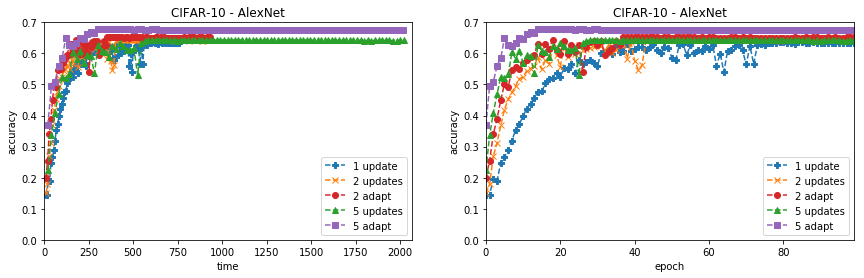}

		\small{Test loss} 
		\includegraphics[width=1.0\textwidth]{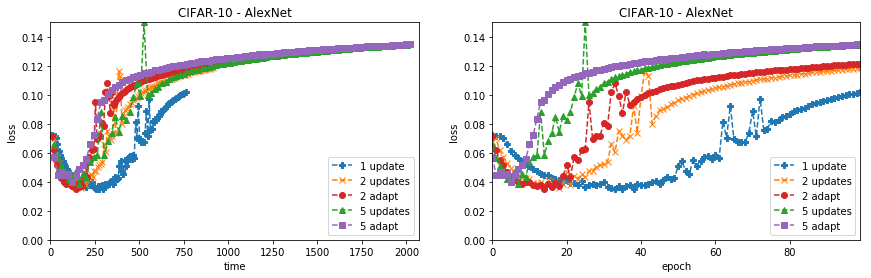}     

 	\caption{Results with minibatch size $m=32$, with and without adaptive learning rate.}                 
	\label{fig:alexnet_32_adaptive} 
	
	~\\
	~\\
  \centering          
  	\small{Test accuracy} 
 		\includegraphics[width=1.0\textwidth]{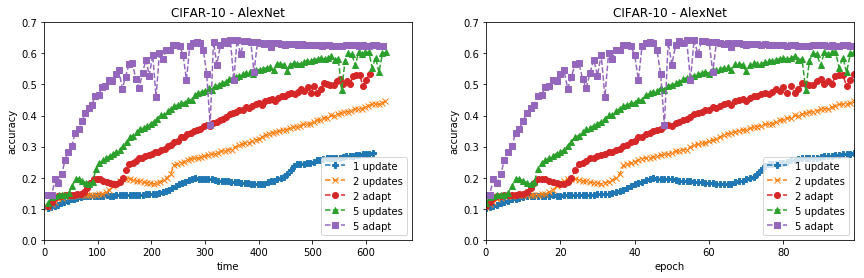}

		\small{Test loss} 
		\includegraphics[width=1.0\textwidth]{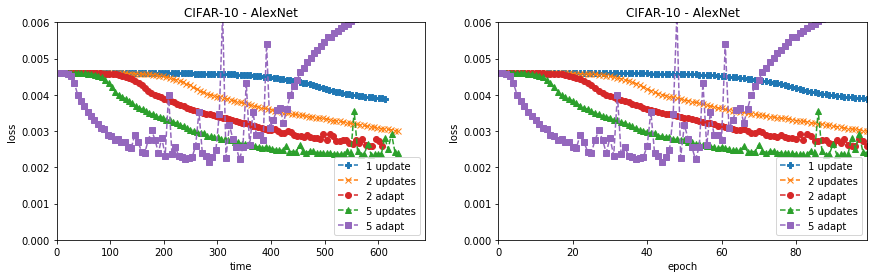}     

 	\caption{Results with minibatch size $m=512$, with and without adaptive learning rate.}                 
	\label{fig:alexnet_512_adaptive}
\end{figure}

\section{Conclusions and future work} \label{sec:conclusions}   

We have investigated the idea of reusing, for $K$ consecutive times, a same minibatch during SGD training. Computational experiments indicate that the idea is promising, at least in the specific setting we considered (AlexNet on CIFAR-10). More extensive tests should be performed on different datasets and neural networks, as one may expect different (possibly worse) behaviors for different settings.

In our tests, for the sake of comparison we fixed a same learning rate $\mu=0.001$ for all values of $K$. However, we also did some preliminary experiments with a different policy that increases the learning rate when reusing the same minibatch, i.e., a learning rate equal to $k \cdot  \mu$ is applied for the $k$-th use of a same minibatch ($k=1,\dots,K$). The rational is that, each time the minibatch is reused, the computed update direction becomes more reliable (at least, for the purpose of optimizing the loss function within the current minibatch). This \emph{adaptive} learning-rate strategy is reminiscent of the Cyclical Learning Rate training policy introduced in \cite{Smith2017}, but it is tailored for our setting. According to the preliminary results reported in Figures \ref{fig:alexnet_32_adaptive}  and \ref{fig:alexnet_512_adaptive}, the approach seems rather effective---a similar behavior has been observed for minibatch size $m=256$ (not reported). Future work should therefore be devoted to better investigate this (or similar) adaptive strategies.

\newpage
\bibliographystyle{plain}  
\bibliography{nips2018_persistency}

\end{document}